\documentclass[journal]{IEEEtran}

\usepackage{multirow}

\ifCLASSINFOpdf
   \usepackage[pdftex]{graphicx}
\else
\fi
\usepackage{array}
\begin{document}

%
\title{An Overview of Color Transfer and Style Transfer for Images and Videos }
%
%
%

\author{Shiguang Liu
\thanks{Shiguang Liu is with College of Intelligence and Computing, Tianjin university, Tianjin 300350, P.R. China.
 (e-mail: lsg@tju.edu.cn).}
}

%
%

\markboth{}%
{Shell \MakeLowercase{\textit{et al.}}: Bare Demo of IEEEtran.cls for IEEE Journals}
%



\maketitle


\begin{abstract}
Image or video appearance features (e.g., color, texture, tone, illumination, and so on) reflect one's visual perception and direct impression of an image or video. Given a source image (video) and a target image (video), the image (video) color transfer technique aims to process the color of the source image or video (note that the source image or video is also referred to the reference image or video in some literature) to make it look like that of the target image or video, i.e., transferring the appearance of the target image or video to that of the source image or video, which can thereby change one's perception of the source image or video. As an extension of color transfer, style transfer refers to rendering the content of a target image or video in the style of an artist with either a style sample or a set of images through a style transfer model. As an emerging field, the study of style transfer has attracted the attention of a large number of researchers. After decades of development, it has become a highly interdisciplinary research with a variety of artistic expression styles can be achieved. This paper provides an overview of color transfer and style transfer methods over the past years. 

\end{abstract}

\begin{IEEEkeywords}
Color transfer, Style transfer, Emotion, Image editing, Video editing.
\end{IEEEkeywords}

%
\IEEEpeerreviewmaketitle

\section{Introduction}
%
%
%
%
\IEEEPARstart{W}ith the prevalence of digital cameras and mobile phones, the  number of images is growing explosively. Image editing has become an important research topic. Various image editing method have been developed, including image adjustment, image enhancement \cite{Liu2018ICPR, Liu2019enhancement, Zhang2016Asia, Zhang2017CAD}, image stitching \cite{Chai2016, Liu2019Stitching}, gamut mapping \cite{Liu2018Gamut}, image recognition \cite{Chen20120ISVC}, image collage \cite{Liu2017Trcollage}, etc. Since color and style are important components of an image, color transfer and style transfer received much attention in recent years.  

Color transfer refers to the process of adjusting the color of an image or a video according to the color of another reference image or video, so that the target image or video has visual color features similar to the source image or video. For example, one can use the method of color transfer to convert a red area in the target image into a blue one. After transferring, the target image or video will show a color style similar to the source image. Another example is to convert images taken in the morning into night style images according to the user's need. Color transfer is often used in image or video processing, image correction, image enhancement and other fields.

The early color transfer method was mainly adjusted manually by users. This processing method need high time complexity and a large amount of user interaction. In 2001, Reinhard et al. \cite{Reinhard2001} pioneered a fully automatic image color transfer technique. Given a target image and a source image, this method can automatically transfer the color of a source image to the target image, that is simple to implement and highly efficient. In recent years, many researchers have proposed color transfer methods for images and videos, such as color transfer based on statistical information (e.g., color mean and variance), and user-assisted color transfer methods with scribbles, swatches and so on. Recently, deep learning based color transfer methods emerged by learning a relationship between the source image and the target image using a deep learning model. As for video color transfer, a direct method is to process a video frame by frame using image transfer techniques. However, temporal inconsistency leads to poor transfer results. By taking into account temporal coherence between neighboring frames, some video color transfer methods also have been developed in recent years. Image or video color is usually related to the emotion the image or video conveys. It is general to convey emotion using colors. Emotion transfer associates the semantic features of images with emotion. By adjusting the color distribution of the source image or video, the emotional color transfer can convert the emotion of a source image or video into the emotion of a target image or video. The existing emotional color transfer methods include color combination based methods and deep learning based methods.

Image and video style transfer technology is to transfer the "style" of a source image (e.g., a painting) to a target image or video, so that the target image or video has the same style as the source image. As the extension of image and video color transfer techniques, image and video style transfer methods also have wide applications. For example, it can transform one's photographing or landscape photos into images with the artistic style of \textit{Monet} and \textit{Van Gogh} and other painters.

Before 2015, researchers usually model a specific texture model to represent a style. Because this modeling method is non-parametric, it requires professional researchers to work manually, which is time-consuming and low-efficient. Moreover, each style model is independent of each other, restricting it from practical applications. Gatys et al. \cite{Gatys2016} began to perform image style transfer by leveraging deep learning networks. In recent years, researchers improved this method and proposed more image style transfers networks to achieve efficient, flexible, and high-quality style results. This book summarized recent progress in image style transfer from the above three aspects. A video style transfer method transforms a style image (e.g., a van Gogh's painting) of a source image to a whole target video sequence. This is a painstaking task if one performs this process by manual. Instead, researchers regard video style transfer as an optimization minimization problem or a learning problem. Therefore, some optimization minimization frameworks and deep learning based frameworks emerged to achieve high definition video style transfer effects over the past few years. 

\section{Image Color Transfer}

Color is one of the most important visual information of an image. Color transfer, a widely used technique in digital image processing, refers to transferring the color distribution of a source image to another target image, so that the edited target image can have a color distribution similar to the source image. The existing color transfer methods can be classified into three categories, namely color transfer methods based on statistical information \cite{Reinhard2001}, color transfer methods based on geometry \cite{Hacohen2011} , and color transfer methods based on user interaction \cite{An2010}.

\subsection{Color Transfer Based on Statistical Information}

\begin{figure}[t] 
\centering
  \includegraphics[width=3.5in]{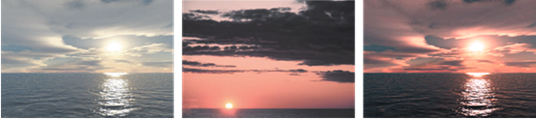}
\caption{Color transfer between images \cite{Reinhard2001}. From left to right: the source image, the target image, and the transferred result. }
\label{ch02.fig1} 
\end{figure}

\textbf{Global Color Transfer.}  In 2001, Reinhard et al. \cite{Reinhard2001} pioneered the technique of image color transfer between images. Figure \ref{ch02.fig1} is a result of this method. Given a target image and a source image, with two statistical indexes (i.e., mean and standard deviation) of an image, this method simply and efficiently transfers the global color statistical information from the target image to the source image, so that the source image has a color distribution similar to the target image, that is, the two images have a similar visual appearance. This method is performed in the $l\alpha\beta$ color space due to the independence between three color channels as follows:

\begin{equation}
{l^{'}}= \frac {\sigma_{t}^{l}}{\sigma_{s}^{l}} (l-\left \langle {l} \right \rangle), \\ 
{\alpha^{'}}=\frac{\sigma_t^\alpha}{\sigma_s^\alpha} (\alpha-\left \langle \alpha \right \rangle), \\ 
{\beta^{'}}=\frac{\sigma_t^\beta}{\sigma_s^\beta} (\beta-\left \langle \beta \right \rangle),
\end{equation}
where $l$, $\alpha$, and $\beta$ are the color components of each pixel in the target image, and  $l^{'}$, $\alpha^{'}$, and $\beta^{'}$ the color components of each pixel in the resulting image, respectively. The $\left \langle l \right \rangle$, $\left \langle \alpha \right \rangle$, and $\left \langle \beta \right \rangle$ are the mean values of $l$, $\alpha$, and $\beta$. The ${\sigma_s^l}$, ${\sigma_s^\alpha} $, and ${\sigma_s^\beta} $ are the standard deviation of the source image and ${\sigma_t^l}$, ${\sigma_t^\alpha} $, and ${\sigma_t^\beta} $ are the standard deviation of the target image, respectively. This algorithm needs to calculate the mean and standard deviation of all three channels of the two images, in which the mean is used to represent the color distribution characteristics of the whole image, and the local details of the image are represented by the standard deviation. As this method treats all pixels in a global manner, it belongs to \textit{global} color transfer techniques.

Following Reinhard et al.'s idea, many impressive image color transfer methods emerged. Piti\'{e} et al. \cite{Pitie2007} proposed a new global color transfer method, which can migrate the color probability density function of any target image to another source image. Pouli and Reinhard \cite{Pouli2011} proposed a \textit{progressive} color transfer approach the histograms for a series of consecutive scales. This method can robustly transfer the color palette between images, allowing one to control the amount of color matching in an intuitive manner. However, when there is a large difference in color distribution between the source image and the target image, the effect of the above global color transfer method may be less satisfactory. To this end, \textit {local} color transfer methods were developed. 

\begin{figure}[hbt] 
\centering
  \includegraphics[width=3.5in]{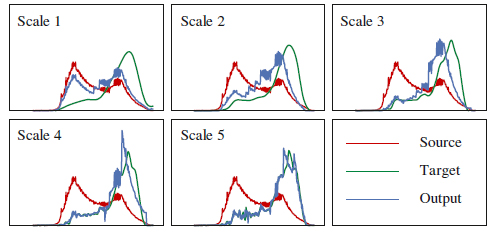}
\caption{Illustration of the histograms for different scales. Note that scale 1 is the coarsest while scale 5 is the finest. The source histogram can progressively approximate that of the target \cite{Pouli2011}. }
\label{ch02.fig9} 
\end{figure}

\begin{figure}[t] 
\centering
  \includegraphics[width=3.0in]{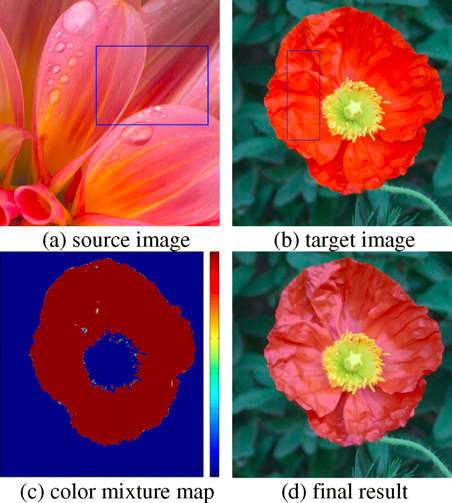}
\caption{The selective color transferring result of a flower scene \cite{Liu2012}.}
\label{ch02.fig2} 
\end{figure}

\textbf{Local Color Transfer.}  Tai et al. \cite{Tai2007} used the Gaussian mixture model to realize local color transfer between images. In this method, the expectation maximization method is exploited to improve the accuracy of color cluster distribution. Later, Irony et al. \cite{Irony2005} proposed a local color transfer method based on higher-level features. This method transfers the color information to each pixel of the target gray image by using a segmented reference image. Nguyen et al. \cite{Nguyen2014} proposed a color transfer method considering color gamut, which can control the color gamut of the transfer result within the range of the color gamut of the source image, making it more consistent with the color distribution of the source image. Hwang et al. \cite{Hwang2014} presented a color transfer method based on moving least squares. This method can produce robust results for different images caused by different camera parameters, different shooting time or different lighting conditions. However, because a large number of feature points are needed as control points, this method requires that the target image and source image share the same scene, which limits the its practical applications. Liu et al. \cite{Liu2012} presented a selective color transfer method between images via an ellipsoid color mixture map. In this method, ellipsoid hulls are employed to represent color statistics of the images. This method computes the ellipsoid hulls of the source and target images and produces a color mixture map to determine the blending weight of pixels in the output image, according to the color and distance information instead of using image segmentation. By mixing the images using the color mixture map, high-quality local color transfer results can be generated (See Figure \ref{ch02.fig2}).

The method of color transfer can be carried out not only between two color images, but also between color images and gray images. The latter is also called gray image colorization. We will discuss this case in the next chapter.

Although the color transfer technique can successfully change the appearance of an image (e.g., changing an image from sunny to cloudy), these methods cannot generate new contents (e.g., new leaves on a dead tree) in the image. When users need to create new contents in the target image, the color transfer method may fail.

\subsection{Color Transfer Based on User Interaction}

In the process of digital image processing, professional software such as Photoshop can help users edit the image appearance flexibly and effectively (e.g., adjusting color, brightness, fusion texture, etc.). However, non-professional users lack skills to operate such softwares. For the convenience of users, stroke based techniques have been introduced into many image editing fields (e.g., color transfer, image segmentation, etc.).

Luan et al. \cite{Luan2007} used a brush based interactive system for image color transfer. This method needs to mark the pixels of all the areas for color transfer with a brush in the image. Wen et al. \cite{Wen2008} leveraged stroke based editing and propagation to develop an interactive system for image color transfer. In comparison with Luan et al.'s approach \cite{Luan2007}, with this method users can get ideal results through a smaller amount of stroke inputs. However, this method cannot deal with the transfer of discrete unconnected regions. 

In recent years, stroke based editing propagation technique has been widely studied. The information that needs to be edited by the user will be automatically propagated to other similar image editing areas with the stroke. To the best of our knowledge, Levin et al. \cite{Levin2004} proposed the first editing propagation framework, which can complete more accurate gray image colorization free of any image segmentation or region tracking. Editing propagation follows a simple criterion: pixels that are spatially adjacent and have similar attributes should have similar colors after editing. Therefore, the color transfer problem is formulated as a global optimization problem. The strokes with editing parameters are used to mark the corresponding regions on the gray image. Based on the above criteria, the marking operation on the strokes is propagated to the adjacent similar regions in the image. The method of editing propagation provides users with a simple and efficient way to edit color of an image. Later, some researchers used global editing propagation techniques to propagate the editing operation on the stroke to the similar areas in the whole image. 

\begin{figure}[hbt] 
\centering
  \includegraphics[width=3.5in]{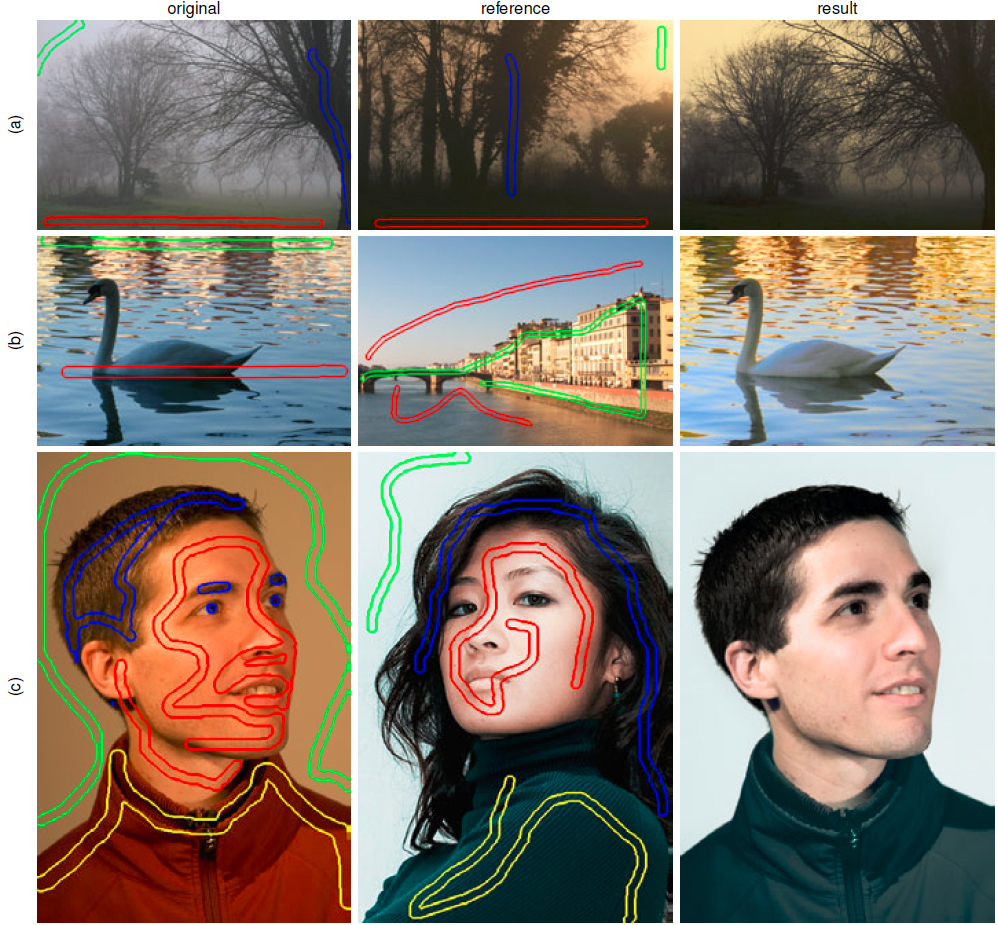}
\caption{Color transfer results on three image pairs \cite{An2010}. From left to right: the source image, the target image, and the transferred result. Note that the strokes in different colors indicate different regions in the corresponding image pairs.}
\label{ch02.fig3} 
\end{figure}

In the above editing propagation methods, the editing information on the strokes needs to be set through parameters. However, in many cases, it is difficult for users to tweak these parameters to meet a desired editing goal. To this end, An et al. \cite{An2010} introduced the reference image in the editing propagation to provide general editing information, so that users can directly obtain the editing parameters of the image by marking the stroke on the region of interest of the color reference image, and then use the editing propagation method to spread the required editing to the whole target image. Figure \ref{ch02.fig3} shows a color transfer result by An et al.'s editing propagation approach \cite{An2010}. Zhang et al. \cite{Zhang2020Access} applied color transfer to a creative design system for creating museum art derivatives. 

However, the existing editing propagation methods assume that the user's stroke marking is accurate, i.e., the marked stroke only covers the pixels of the user's region of interest. When the stroke mark does not meet this assumption, it will directly affect the final editing and propagation results. Therefore, Subr et al. \cite{Subr2013} and Sener et al. \cite{Sener2012} used Markov random fields to solve this problem. However, the introduction of the Markov random fields leads to a high computational complexity, which limits its practical application.

This book focuses on color transfer between images. However, except color, more features (e.g., texture, contrast, and tone) contribute to the appearance of an image. More generally, given a relationship between two images, it can be
replicated on a new image by the technique of image analogies \cite{Hertzmann2001}.

\subsection{Hybrid Transfer}

\begin{figure}[t] 
\centering
  \includegraphics[width=3.5in]{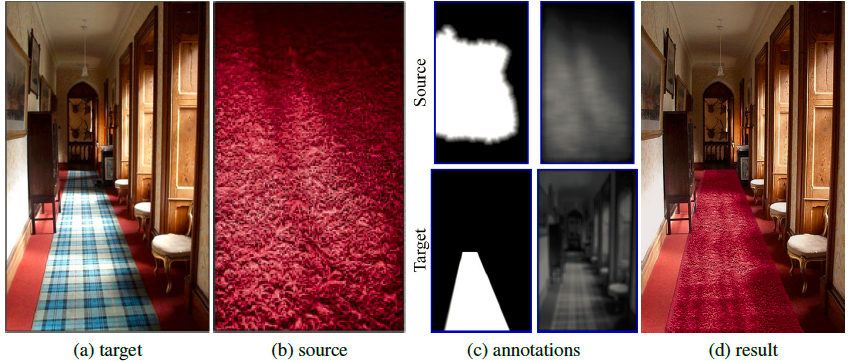}
\caption{Replacing the carpet in an image by the texture transfer technique \cite{Diamanti2015}. Given a target image (a) and a source texture image (b), the source and target are annotated according to lighting amount and fine-scale carpet fiber transitions (c), and a new carpet (d) is finally produced via the texture transfer technique.}
\label{ch02.fig4} 
\end{figure}

\textit{Image Texture Transfer.} Texture is another critical image feature. In recent years, texture synthesis technique, especially the sample based texture synthesis method, has made remarkable progress \cite{Wei2009}. The core idea of sample based texture synthesis method is to generate a new texture similar to the sample according to the surface features of a given texture sample. Image analogy \cite{Hertzmann2001}, also known as texture transfer \cite{Efros2015}, refers to synthesizing new textures in a target image to make the target image and the reference image share similar texture appearance. Bonneel et al. \cite{Bonneel2010} used texture transfer to enhance the appearance of 3D models. With a photograph as the source image, a series of computer synthesized target images become more realistic by using color and texture transfer to improve the roughness of the appearance of the target images \cite{Johnson2010}. Zhang and Liu \cite{Zhang2011} addressed texture transfer in frequency domain. This algorithm is efficient due to various FFT (Fast Fourier Transform) algorithms for transforming images to frequency domain, that can utilize any of the texture synthesis algorithms to generate a texture of the proper size, and then transfer this texture to the target image. A set of tools and parameters are also provided to control the process of this method. Diamanti et al. \cite{Diamanti2015} manually annotated the sample image to mark the areas that need to generate texture in the process of texture synthesis. As shown in Fig. \ref{ch02.fig4}, the carpet in the target image is successfully replaced with Diamanti et al.'s texture transfer method. Although texture transfer can synthesize new content in a target image, it often covers the structural details that should be retained in the target image.

Shih et al. \cite{Shih2013} used the image of a scene at one time to predict the appearance of the scene at another time. This method can also realize the prediction of the corresponding night image by taking the daytime image as the source image. However, if the night image is the source image, because only color transfer is considered, the details of the dark places at night will be lost, and it then fails to predict the corresponding daytime image. 

\textit{Image Tone Transfer.} The pixel colors may vary greatly in one image due to different lighting environments, leading to distinct visual discontinuities. To this end, Liu et al. \cite{Liu2011Toning} proposed multi-toning image adjustment method via color tansferring. An unsupervised method is used to cluster the source and the target image based on color statistics. A correspondence is established via matching the texture and luminance between the subsets in the source and target images. The color is then transferred from the matched pixels in the source to the target. Finally, the graph cut is utilized to optimize the seams between different subsets in the target image. 

Bae et al. \cite{Bae2006} proposed to transfer the tone from a source photo to make a target image have a photo graphics look. Based on a two-scale non-linear decomposition, this method modifies different layers of an image via their
histograms. A Poisson correction mechanism is designed to prevent potential gradient reversal and preserve image detail.

Without a source image, the tone of a given can also be creatively adjusted by a user.  For example, Lischinski et al. \cite{Lischinski2006} proposed to adjust the tone an image in the regions of interest with strokes. To achieve a finer tone control, Farbman et al. \cite{Farbman2008} manipulated the tone of an image from the coarsest to the finest level of detail with a multi-scale decomposition of the image.  

Tone reproduction for dynamic range reduction also gains much interest from researchers. Various operators (e.g., logarithms, histograms, sigmoids, etc.) have been devised for tone curves \cite{Drago2003, Ward1997, Reinhard2002, Durand2002, Fattal2002, Li2005}.

The above color transfer and texture transfer methods can process the image appearance from the perspective of a single feature, i.e., color or texture, so that the target image is similar to the source image in terms of color or texture. However, the color and texture of an image cannot be changed at the same time by using color transfer or texture transfer alone. 

\begin{figure}[hbt] 
\centering
  \includegraphics[width=3.5in]{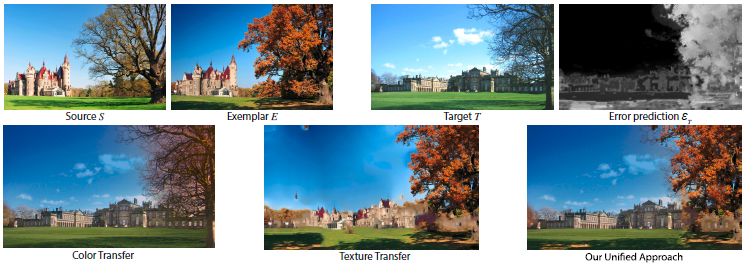}
\caption{Results generated with the hybrid transfer method \cite{Okura2015}. Given source, exemplar, and target images, this method automatically transfers leaves in the trees. Note that color transfer does not change the texture on the tree in the target, while texture transfer destroys the target structure.}
\label{ch02.fig5} 
\end{figure}

\textit{Image Hybrid Transfer.} In order to tackle this challenge, Okura et al. \cite{Okura2015} first proposed an image hybrid transfer method by combining color transfer and texture transfer. This method can take full advantages of the two transfer methods. It can not only change the color of the target image, but also can generate new content in the target image. Figure \ref{ch02.fig5} shows an experimental result of this method. This method uses a ternary image group as input: a target image, a sample image, and a source image. The target image and sample image are required to be photos of the \textit{same} scene taken at different times, and need to have the same perspective. The source image needs to have some visual similarity with the target image. By predicting the probability of a successful color transfer between the sample image and the source image, this method calculates the error between the target image and the sample image, so as to judge the areas requiring color transfer and texture transfer in the target image. Among them, the region with small error is processed by color transfer; For regions with large errors, texture transfer is required. This method primarily uses the fact that the sample image is the real transfer result of the target image to predict and process the target image, and finally make the target image have the appearance similar to the sample image. However, this method suffers from two limitations. On one hand, the requirements for the input image are relatively strict, which limits its practical application to a great extent; On the other hand, limited by the methods based on color transfer, texture transfer and image matching, inaccurate matching and large errors may occur in the results of color transfer and texture transfer, and therefore fail to generate an ideal image transfer result.

Song and Liu \cite{Song2017} presented a novel image hybrid transfer method combining color and texture features. The algorithm does not require the user to provide a ternary image group. Instead, it mainly relies on feature extraction and feature matching to automatically judge the regions that need texture transfer or color transfer in the source and target images. This method allows users to mark the areas to be transferred using strokes. Color transfer and texture transfer are then performed the transferring task between corresponding regions. Later, Liu and Song \cite{Liu2018AppearanceTansfer} extended the above method to dealing with image appearance with multi-source images.

Albeit successful, the hybrid transfer methods may fail if there are large illumination differences between the source image and the target image. Especially when there is strong illumination in the image, the color deviation caused by illumination or the light that can be seen in the target image will affect the result of color transfer. Because the traditional color transfer method does not consider the problems caused by these lighting factors, it cannot produce results similar to the source image. To this end, Liu et al. \cite{Liu2019} devised an appearance transfer method between images with complex illumination images via deep learning. We will elaborate this method in the next section.

\subsection{Deep Learning Based Color Transfer}
Deep learning based methods aim to transform the transfer problem into a nonlinear regression problem, and then solve the regression problem through the powerful ability of deep learning method, so as to obtain an appearance mapping relationship between the source image and the target image. With the trained network model, users can apply the learned specific mapping relationship to an input target image to obtain a similar appearance transfer result.


He et al. \cite{He2017} proposed a neural color transfer between images. They use neural representations (e.g., convolutional features of pre-trained VGG-19) to achieve dense semantically-meaningful matching between the source and target images. This method is free of undesired distortions in edges or detailed patterns by performing local color transfer in the image domain instead of the feature space. 

Liao et al. \cite{Liao2017} proposed "deep image analogy" to achieve visual attribute transfer (e.g., color, texture, tone, and style) across images with very different appearance but perceptually
similar semantic structure. Aiming at better representations for semantic structures, a deep Convolutional Neutral Network (CNN) \cite{Krizhevsky2017} was employed to construct a feature space in which to form image analogies.

Lee and Lee \cite{Lee2016} transferred the appearance of noon images to night images using CNN. Liu et al. \cite{Liu2019} designed a deep transfer framework for images with complex illumination. This method consists of illumination transfer and color transfer. CNN is employed to extract the hierarchical feature maps. of which the former is carried out in the illumination channel, while for the other two channels, the latter is performed to transfer the color distribution to the target image. This method can achieve progressive transfer using the histogram reshaping method with hierarchical feature maps extracted from the CNN. The final appearance transfer results are output by combining the illumination transfer and color transfer. Finally, the joint bilateral filter is adopted to smooth the noises so as to further optimize the transfer results.

Terry Johnson from UK pointed out that image color transfer can change the brightness of an image in two ways. The subject image can be modified to adopt the luminance characteristics of the reference image but over and above this, and the luminance can be indirectly modified by any change in the chromatic content of the subject image \cite{Terry2022}. He also mentioned that the deep learning based color transfer method is dependent upon the choice of images for the training set and performed some tests \cite{Terry2022CT}.


\section{Emotional Color Transfer}

Color is widely used to represent the emotion of an image. The emotion of a target image can be changed by adjusting its color distribution. Emotional color transfer aims to transfer the emotion of the source image to that of the target image by changing the color distribution of the target image. Emotional transfer can be widely used in multimedia processing, advertising, and movie special effects. For example, through the technique of emotional color transfer, one can transfer a given image into a brighter and fresher one with a happy emotion. This technique has great potential to inspire designers and improve their work efficiency. The existing emotion transfer methods can be classified into two categories, emotional color transfer method based on color combination and emotional color transfer method based on deep learning. 

\subsection{Color Combination Based Emotional Color Transfer}

The emotional color transfer method based on color combination mainly considers the low-level semantic information, i.e., color of an image, and then transfers the color distribution of the image to the target image, so that the color distribution of the target image is consistent with that of the source image. 

\subsubsection{Emotion Transfer Based on Histogram}

\begin{figure}[t] 
\centering
  \includegraphics[width=3.5in]{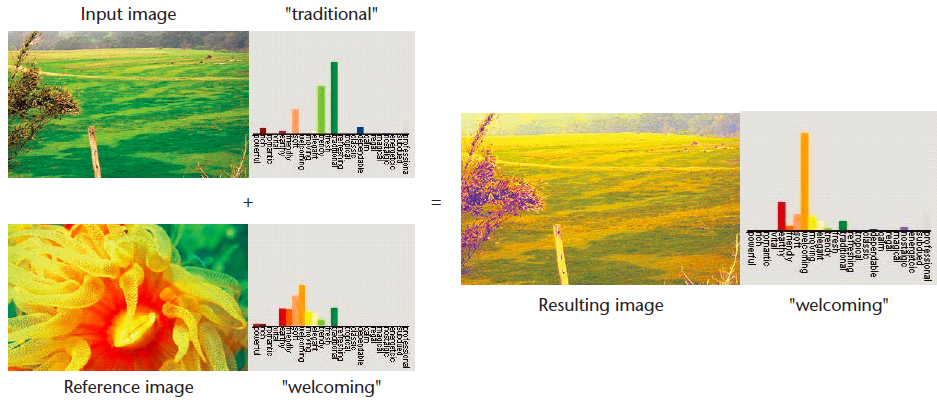}
\caption{A color-mood transfer result \cite{Yang2008}.}
\label{ch03.fig1} 
\end{figure}

Yang and Peng \cite{Yang2008} considered the relationship between image emotion and color in color transfer and used histogram matching to maintain spatial consistency between image colors. This method uses the $RGB$ color space for image classification and color transformation. In order to align the target image more closely with the source image, Xiao and Ma's local color transfer method \cite{Xiao2006} is employed to match the color distribution of the target image and the source image. The PCA (Principal Component Analysis) technique is used to adjust the main trend or distribution axis of the two images in the $RGB$ space, and then match the distribution of the target image with that of the source image. Specifically, this method is performed as follows: Firstly, the covariance matrix of pixel values involved in the $RGB$ space is summed; Then, the eigenvalues and eigenvectors of the matrix are determined. The eigenvector corresponding to the maximum eigenvalue represents the desired principal axis, which can be used to derive the rotation transformation of aligning the two color distributions. Finally, the adjusted $RGB$ values are cut to a normal range. Figure \ref{ch03.fig1} shows an emotional color transfer result of Yang and Peng's method \cite{Yang2008}. 

\subsubsection{Emotion Transfer Based on Emotion Word}

Given an emotional word (e.g., enjoyable, pretty, heavy, etc.), Wang et al. \cite{Wang2013} proposed an emotional word based emotional color transfer method between images. They introduced color theme and devised the emotional word relationship model to select the best color theme from the candidate themes. 

\begin{figure}[t] 
\centering
  \includegraphics[width=3.5in]{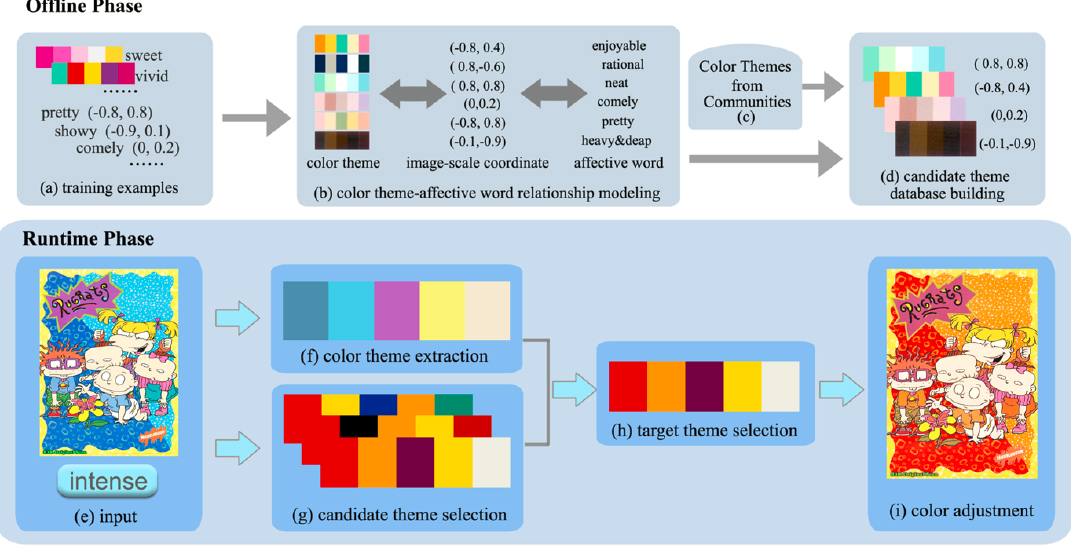}
\caption{Illustration of the pipeline of the emotional color transfer framework based on emotional word. Note that this framework is composed of the offline stage and the runtime stage. In the offline phase, the color theme-affective word relationship (b) is constructed from the training examples (a). The candidate theme database (d) is built based on this relationship model and on-line communities (c). In the runtime stage, given the input image and an affective word (e), candidate themes (g) are chosen with the affective word from the candidate theme database (d). By matching the candidate themes with the color theme (f), most suitable target theme (h) is selected and finally the target image (i) is edited according to the target theme \cite{Wang2013}.}
\label{ch03.fig2} 
\end{figure}

This emotion transfer algorithm is automatic and requires no user interaction. Figure \ref{ch03.fig2} is the system framework, which is composed of an offline stage and a running stage. In the offline stage, the relationship between the color theme and an emotional word is constructed by using the art theory and empirical concepts \cite{Arnheim1954, Itten1974, Kobayashi1995, Matsuda1995}. Emotional words and color themes are mapped to an image scale space. The 180 emotional words in the Kobayashi theory \cite{Kobayashi1995} have predefined image scale coordinates that are regarded as marker words. Here, the 5-color theme is employed because it can fully represent images with rich colors. By using artistic principles, The CC features (color composition theory guided feature descriptors) are designed, by which the color theme model with the image scale space are compared. Finally, Internet resources are exploited to expand the database to more than 400,000 color themes. In the running stage, given an emotional word, the standard semantic similarity is computed to obtain the weight corresponding to each marker word, and get the coordinates in the image scale space. The closest candidate color theme is selected. Then, a strategy is designed to choose the most suitable color theme, balance the appearance compatibility with the target image, and adapt to the emotional word. Finally, the RBF (Radial Basis Function) based interpolation method is utilized to adjust the image appearance according to the selected color theme.

\subsubsection{Emotion Transfer Based on Color Combination}

\textbf{Color Combination Extraction.} Color combination of an image, also known as color theme, refers to the color set that can reflect the main colors and color matching information in an image. The color theme extraction methods of images can be divided into three categories: clustering segmentation methods, statistical histogram methods, and data driven methods.

\textit{The clustering segmentation method} is a common color theme extraction method. An image is spatially clustered to obtain a locally clustered block image, so as to extract the colors representing each block respectively. The K-means \cite{Means1999} and C-means \cite{Hathaway2001} are the most used methods for clustering. The K-means method introduces the clustering number $k$, initially randomly selects $K$ pixels as the clustering center, and iteratively transforms the clustering center to divide the image into $k$ parts by minimizing the overall difference. The K-means method ignores the small color areas in an image, but one's favor for color is not ignored because of the proportion. The C-means method is similar to the K-means method, except that when minimizing the overall difference, it introduces a judgment on individual outliers to determine whether they belong to a new class or just individual outliers. The C-means method does not avoid the shortcomings of the K-means method in color theme extraction. Image segmentation methods \cite{Luo2014} have also been applied to the extraction of color themes. Wang et al. \cite{Wang2013} used the graph segmentation to divide an image into a small number of blocks. They scan and record the average pixel value of each block, and then merge the blocks with similar pixel values, so as to achieve the required number of colors of a color theme.

\textit{The statistical histogram method} stems from the distribution of pixel values in an image. Apart from spatial factors, only the number of occurrences of pixel values in an image is counted. The interval or point of pixel value aggregation in an image is obtained through statistical analysis as the color theme of an image. Delon et al. \cite{Delon2005} converts an image into the HSV color space \cite{Chen2008}, counts the distribution histogram of $H$ component, and divides the histogram into a region. The criterion of division is to divide according to the minimum value. Then, the region of $H$ distribution is obtained respectively, the $S$ and $V$ components are separately counted, so as to generate the main color of the given image. Morse et al. \cite{Morse2007} controls a statistical amount when dividing, so as to reduce the number of colors. Zheng et al. \cite{Zheng2012} transformed the color space into three new emotional spaces and made histogram statistics according to the three emotional spaces. These methods count the color values that appear the most times. However, the theme one feels for an image cannot be simply determined by the number of occurrences.

\textit{The data driven method} is another commonly used color theme extraction method. Firstly, the algorithms mentioned in the above two categories are used to extract the color theme, and then the classifier is employed for optimization. Donovan et al. \cite{Donovan2011} extracts the color theme of an image by using the DERIECT algorithm \cite{Johnson1993}. On the other hand, according to the ranking of the themes on the image website, for the top color themes, the first-order linear LASSO regression model \cite{Tibshirani1996} is used to select important features. When optimizing the color theme, they select a series of similar color themes, and rely on the ranking of the color theme in the database when determining the color theme. This method maintains the quality of color theme well, and the extracted color theme has high aesthetic value. However, this method pays too much attention to the original ranking of color themes and ignores one's preference for the connotation in an image. In addition, the color theme trained by this method suffers from over saturation, and the visual sense of color theme is less satisfactory. 

\begin{figure}[t] 
\centering
  \includegraphics[width=3.5in]{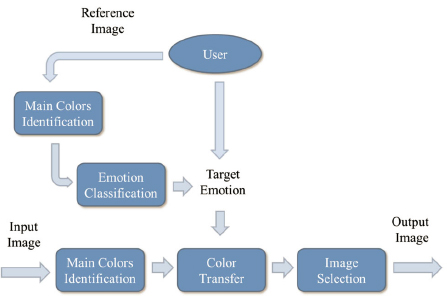}
\caption{The overview of the emotional color transfer method based on color combination \cite{He2015}.}
\label{ch03.fig3} 
\end{figure}

Liu et al. \cite{Liu2018theme} presents an approach to extracting color themes from a given fabric image. They use a saliency map to judge the visual attention regions of and separate the image into visual attention and non-visual attention regions. The dominant colors of two regions are respectively computed and then merged to produce an initial target color theme. Without considering the emotion factors, the above color theme extraction methods can only produce a single color theme result for an image, which does not satisfy one's favor on different colors under different mood
states. Liu and Luo \cite{Liu2016, Liu2014} presented a method to extract the emotional color theme from an image. They perform theme extraction with emotion value of each pixel rather than color value. The emotional discrepancy between colors in the theme is used to evaluate a color theme quality. Thus, this method establishes the emotional relationship between the target theme and the candidate theme.

\textbf{Emotional Color Transfer Using Color Combination.} On the basis of the above color theme extraction methods, some researchers developed emotional color transfer approaches between images. He et al. \cite{He2015} proposed an emotional color transfer framework (see Fig. \ref{ch03.fig3}) based on color combination. The framework identifies three colors in an image by using the expectation maximization clustering algorithm, and models the color transfer process as two optimization problems, 1) calculating the target color combination, and 2) maintaining the gradient of the target image. The user can select the target emotion by providing source images or directly selecting emotion keywords. If a source image is available, one can extract the main color (the clustering center in the color space) from the source image and use the closest scheme in the color emotion scheme as the target scheme. In this way, a specific scheme is selected and the color combination in the scheme is used for color transfer.

\begin{figure}[t] 
\centering
  \includegraphics[width=3.5in]{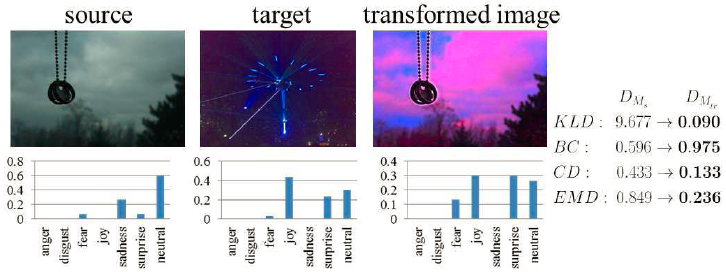}
\caption{An result of transferring evoked emotion distribution by \cite{Peng2015}. The color and texture related features are transformed from the source image to the target image. The real probability distribution of the evoked emotion is indicated under each image, demonstrating that the color adjustment makes the target image more joyful. On the right is a quantitative evaluation of the similarity between the distribution and the four metrics $M \quad (M \in {KLD,\quad BC, \quad CD, \quad EMD})$, where $D_{M_s}$ is the distance between the source and the target distribution, and $D_{M_{tr}}$ denotes the distance between the transferred image and target distribution. For each measure, the better result is shown in bold. It can be observed that the transferred image can endow the target image with more similar emotion compared to that of the source image.}
\label{ch03.fig5} 
\end{figure}

Meanwhile, the main color in the target image is also extracted. Once the main color in the target image is the same as the target scheme, all the colors in the target image is transferred to 24 output images (each scheme has 24 three-color combinations). Then, the best output image in the 24 three-color combinations is selected. Here, the aim of color transfer is to calculate the target color combination.

 
Liu and Pei \cite{Liu2018Access} presented a texture-aware emotional color transfer method between images, that can realize emotion transfer with an emotion word or a source image. Given an emotion word, this method can automatically change an image to the target emotion. Here, an emotion evaluation method is designed to compute the target emotion from a source image. The three color-emotion model databases are constructed to seek the proper color combinations expressing the target emotion. At last, a color transfer algorithm is designed to ensure the color gradient and naturalness by taking advantages of both color adjustment and color blending.


 
\subsection{Deep Learning Based Emotional Color Transfer}

Peng et al. \cite{Peng2015} proposed a novel image emotion transfer framework based on CNN, in which they predict emotion distributions instead of simply predicting a single dominant emotion evoked by an image. Figure \ref{ch03.fig5} shows the change of emotion distribution of a target image by this method.

The emotional color transfer framework first shows that different people have different emotional changes to an image, i.e., one may also have a variety of different emotions for a same image. The framework uses the Emotion6 database to model emotion distribution. The Emotion6's emotion distribution predictor is used as a better baseline than previously used support vector regression (SVR) \cite{Machajdik2010, Solli2010, Wang2013}. The framework also predicts emotions in the traditional environment of emotional image classification, indicating that CNN is superior to Wang et al.'s method \cite{Wang2013} in artphoto dataset. Finally, with the support of a large-scale (600 image) user study, the framework successfully adjusts the emotional distribution of the image to the emotional distribution of the target image without changing the high-level semantics.

\begin{figure}[t] 
\centering
  \includegraphics[width=3.5in]{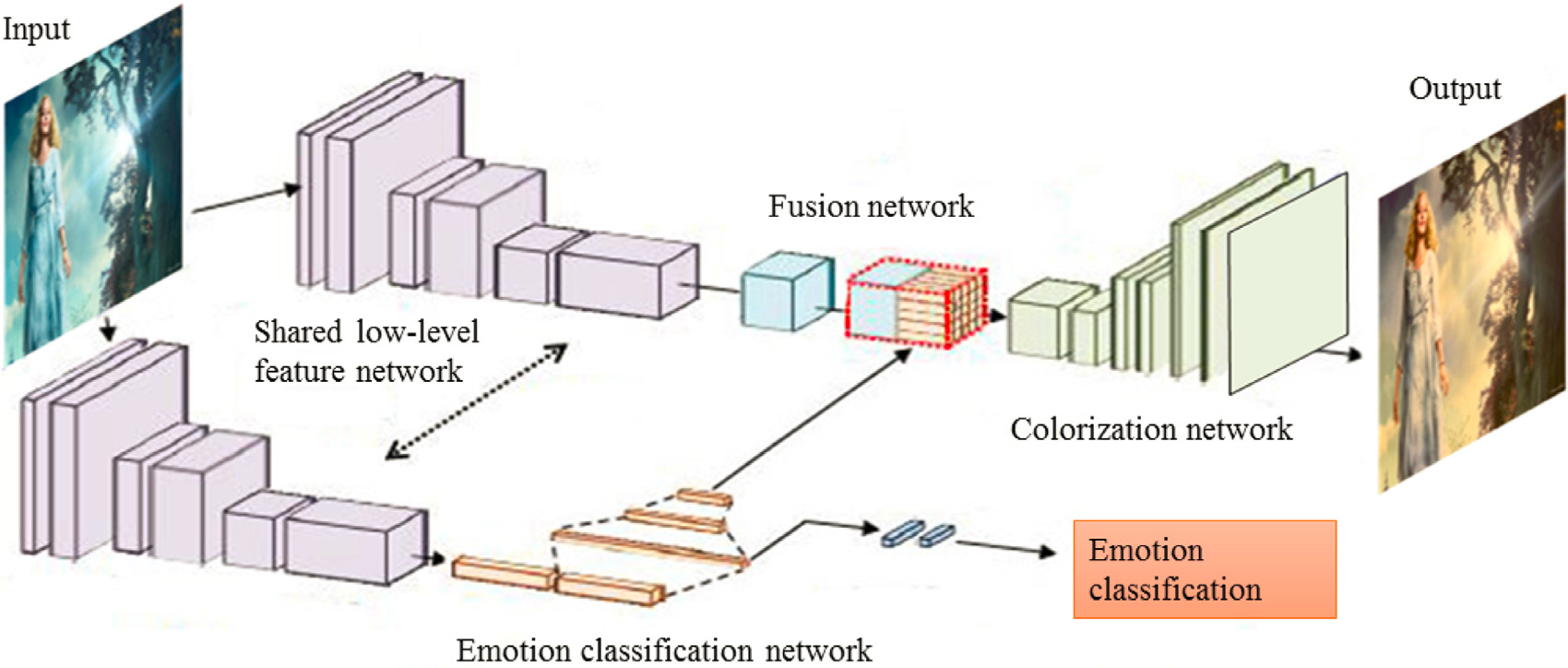}
\caption{The network structure of the emotional color enhancement framework based on deep learning \cite{Liu2018}.}
\label{ch03.fig6} 
\end{figure}

Liu et al. \cite{Liu2018} proposed an emotional image color transfer method via deep learning. This method takes advantages of both the global emotion classification and local semantic for better emotional color enhancement. As shown in Fig. \ref{ch03.fig6}, this method is comprised of four ingredients, a shared low level feature network (LLFN), an emotion classification network (ECN), a fusion network (FN), and a colorization network (CN). The LLFN takes in charge of extracting the semantic information. The ECN serves to constrain the color so that the enhancement results is more consistent with the source emotion. The ECN aims to treat the low-level features with four convolutional layers followed by fully-connected layers. The ECN and the LLFN are then catenated by a fusion network. Finally, the CN is employed to create the emotional transferred results.

\begin{figure}[hbt] 
\centering
  \includegraphics[width=3.5in]{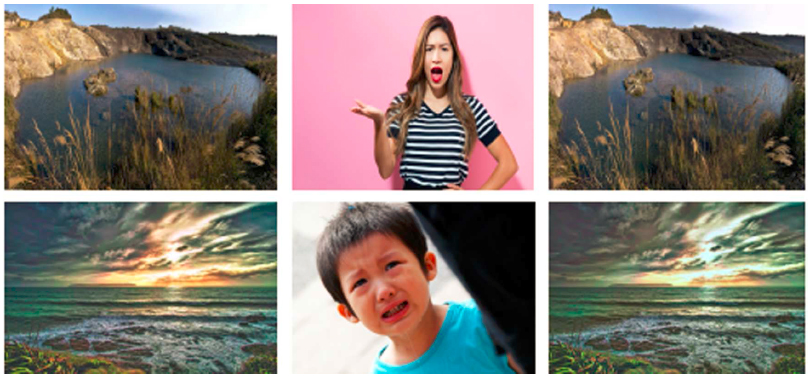}
\caption{Two failure cases of traditional emotional transfer methods. From left to right: the source image, the target image, and the result image.}
\label{ch03.fig7} 
\end{figure}

\textbf{Facial-Expression-Aware Emotion Transfer.} The existing deep learning methods do not consider the impact of face features on image emotion. When the background emotion in an image is inconsistent with the emotion expressed by the face in the image, the existing methods ignore the emotion expressed by the face, which will inevitably lead to the inconsistency between the emotion of the resulting image and the emotion of the source image. For example, if the background of the target image is dark while the facial expression is happiness, previous methods would directly transfer dark color to the source image, neglecting the facial emotion in the image. As shown in Fig. \ref{ch03.fig7}, the facial expression of the target image in the bottom is sad, while the background color is very bright, which is in conflict with each other. 

To this end, \cite{Pei2018, Liu2022} proposed an emotion transfer framework of face images based on CNN. The framework considers the face features of the image and accurately analyzes the face emotion of the image. Given a target face image, this method first predicts the facial emotion label of the image using an emotion classification network. The facial emotion labels are then matched with pre-trained emotional color transfer models. The pre-trained emotional models consists of 7 categories, namely the anger model, the disgust model, the happiness model, the fear model, the sadness model, the surprise model, and the neutral model. For example, if the predicted emotion label is “happy,” then the corresponding emotion model is the happiness model. The matched emotion model is used to transfer the color of the source image to the target image. Different from the previous methods, the emotion transfer model is obtained by a pre-training network. The network can simultaneously learn the global and local features of the image. Additionally, an image emotion database called "face-emotion" is established. Different from the previous emotional image database, the images contained in the database are color images with facial expression features and obvious background, such as green mountains and rivers. The face emotion database provides useful data for facial emotion analysis.


\section{Video Color Transfer}

Either image and image sequence can arouse emotion. In 2004, Wang and Huang \cite{Wang2004} proposed an image sequence color transfer algorithm, that can render an image sequence with color distribution transferred from three source images. The mean and variance of the colors are computed in each source image, which are then interpolated with a color variation curve (CVC) to generate in-between color transfer results. However, this method suffers from a limitation that only three source images can be allowed for interpolation for an in-between image sequence.

\begin{figure}[hbt] 
\centering
  \includegraphics[width=3.0in]{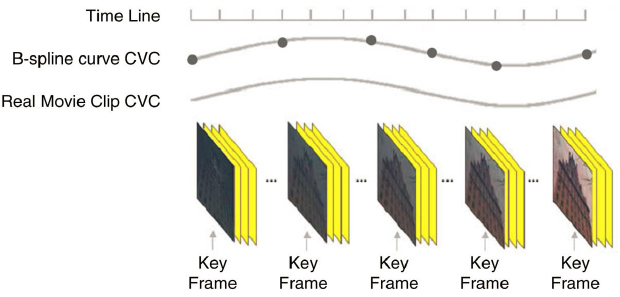}
\caption{Overview of the real movie clip CVC and the B-spline curve CVC \cite{Wang2006}. }
\label{ch02.fig8} 
\end{figure}

Wang et al. \cite{Wang2006} further presented a more general image sequence color transfer method. Given a video clip as the input source, by analyzing the color mood variation with B-spline curves, they create a resulting image sequence from only one image. 
Instead of only supporting a linear or parabolic color variation curve, this method employs a generalized color variation curve (GCVC) for more flexible control for color transfer across in-between images. As shown in Fig. \ref{ch02.fig8}, the B-spline allows the user to more effectively design a curve by the control points, which can achieve similar results with the brute force method. 

In movie productions, artists usually painstakingly tweak the color palette of a film to achieve a desired appearance, which is also known as \textit{color grading}. Bonneel et al. \cite{Bonneel2013} proposed an example based video color grading approach, that automatically transfers the color palette of a source video clip to a target video. Firstly, a per-frame color transform is computed to map the color distribution from the target video to that of the source video. To reduce artifacts (e.g., bleeding and flickering), a differential geometry-based scheme is then devised to interpolate the transformations so that their curvature can be minimized.  Finally, the keyframes that best represent this interpolated transformation curve are determined for the subsequent process of color grading.

Aiming at a full nonlinear and non-parametric color mapping in the 3D color space, Hwang et al. \cite{Hwang2014} introduced moving least squares into color transfer, and proposed a scattered point interpolation framework with a probabilistic modeling of the color transfer in the 3D color space to achieve video color transfer. 

Recently, Liu and Zhang \cite{Liu2021VideoTransfer} proposed a novel temporal-consistency-aware video color transfer method via a quaternion distance metric. They extracted keyframes from the target video and transfer their color from the source image by the Gaussian Mixture Models (GMM) based a soft segmentation algorithm. By matching the pixel through a quaternion-based distance metric, they transfer color from keyframes to the in-between frames in an iterative manner. An abnormal color correction mechanism is further devised to improve the resulting color quality. This method is capable of better preserving the temporal consistency between video frames and alleviating the color artifacts than the state-of-the-art methods.


\section{Conclusion and Future Work}

This paper overviews the methods of color transfer and style transfer for image and videos. We summarized image color transfer approaches based on statistical information and user interaction. Then we discussed the hybrid image hybrid transfer methods by combining different image features such as color and texture. We further introduced the emotional color transfer methods, including histogram based methods, emotion word based methods, and color combination based methods. We further summarized style transfer approaches from three perspectives, i.e., efficiency-aware, flexibility-awre, and quality-aware style transfer methods. Finally, we also introduced the latest deep learning based color transfer and style transfer methods for images and videos.

Until now, much progress has been made in the field of color transfer and style transfer for images and videos. However, there are still some problems to be solved in this field. For an instance, how to further improve the quality of the transferred results according to the user desire is still a challenge. The user also wish to perform color and style editing in real time. So the computational efficiency also needs further improved. Moreover, it is necessary to develop lightweight color and style editing tool on mobile platforms, since users usually need to modify their photos on such devices. It is an interesting topic to build specific metrics by exploiting modern image quality assessment techniques (e.g., \cite{Gao2021, Gao2022, Huang2018MM, Liu2019TOMM}) for color transfer and style transfer. It is another direction to develop color transfer and style transfer techniques for non-photorealistic images (e.g., cartoon and line drawings \cite{Liu2022MTAP, Liu2022FCS, Zhang2019Heritage}).

\ifCLASSOPTIONcaptionsoff
  \newpage
\fi


\begin{thebibliography}{1}




\bibitem{An2010}
X. An, F. Pellacini,
``User-controllable color transfer,''
\textit{Computer Graphics Forum},
vol.~29, no.~2, pp.~263--271, 2010.

\bibitem{Anderson2016}
A. Anderson, C. Berg, D. Mossing, et al.,
DeepMovie: using optical flow and deep
neural networks to stylize movies``
\textit{ArXiv 1409.1556}, 2016.

\bibitem{Arnheim1954}
R. Arnheim,
``\textit{Art and Visual Perception: A Psychology of the Creative Eye},''
Univ. of California Press, 1954.

\bibitem{Bae2006}
S. Bae, S. Paris, and F. Durand,
``Two-scale tone management for photo graphic look,''
\textit{ACM Transactions on Graphics},
vol.~25, no.~3, pp.~637--645, 2006.


\bibitem{Bonneel2010}
N. Bonneel, M.V.D. Panne, S. Lefebvre, et al.,
``Proxy-guided texture synthesis for rendering natural scenes,''
In \textit{Proceedings of Vision, Modeling \& Visualization Workshop},
pp.~87--95, 2010.

\bibitem{Bonneel2013}
N. Bonneel, K. Sunkavalli, S. Paris, and H. Pfister,
``Example-based video color grading,''
\textit{ACM Transactions on Graphics},
vol.~32, no.~4, pp.~39:1--39:12, 2013.

\bibitem{Chai2016}
Q. Chai, and S. Liu,
``Shape-optimizing hybrid warping for image stitchings,''
In \textit{Proceedings of the IEEE International Conference on Multimedia and Expo (ICME )},
Article number: 7552928, pp. 1--6, 2016.

\bibitem{Chen2008}
T.W. Chen, Y.L. Chen, and S.Y. Chien,
``Fast image segmentation based on K-Means clustering with histograms in HSV color space,''
In \textit{Proceedings of the 10th IEEE Workshop on Multimedia Signal Processing},
pp.~322--325, 2008.

\bibitem{Chen2016}
W. Chen, Z. Fu, D. Yang, et al.,
``Single-image depth perception in the wild,''
In \textit{Proceedings of Advances in Neural Information Processing Systems (NIPS)},
pp. 730-–738, 2016.

\bibitem{Chen2017}
D. Chen, L. Yuan, J. Liao, N. Yu, and G. Hua,
``StyleBank: An explicit representation
for neural image style transfer,''
In \textit{IEEE Conference on Computer Vision and Pattern Recognition (CVPR)}, 
pp. 2770-–2779, 2017.

\bibitem{Chen2017ICCV}
D. Chen, J. Liao, L. Yuan, et al.,
``Coherent online video style transfer,''
In \textit{Proceedings of IEEE International Conference on Computer Vision (ICCV)}, 
pp. 1114-–1123, 2017.

\bibitem{Chen20120ISVC}
Y. Chen, and S. Liu,
``Deep partial occlusion facial expression recognition via improved CNN,''
In \textit{Proceedings of International Symposium on Visual Computing (ISVC)},
pp. 451--462, 2020.

\bibitem{Deng2020}
Y. Deng, F. Tang, W. Dong, W. Sun, F. Huang, and C. Xu,
``Arbitrary Style Transfer via Multi-Adaptation Network,''
In \textit{Proceedings of ACM Multimedia},
pp. 2719--2727, 2020.

\bibitem{Deng2021}
Y. Deng, F. Tang, W. Dong, H. Huang, C. Ma, and C. Xu, 
``Style transfer via multi-channel correlation,''
In \textit{Proceedings of The Thirty-Fifth AAAI Conference on Artificial Intelligence (AAAI)},
pp. 1210--1217, 2021.

\bibitem{Delon2005}
J. Delon, A. Desolneux, and J.L. Lisani,
``Automatic color palette,''
In \textit{Proceedings of IEEE International Conference on Image Processing},
pp.~706--709, 2005.

\bibitem{Deng2022}
 Y. Deng, F. Tang, W. Dong, C. Ma, X. Pan, L. Wang, and C. Xu, ``StyTr2Image Style Transfer with Transformers,''
 In \textit{Proceedings of IEEE/CVF Conference on Computer Vision and Pattern Recognition (CVPR)},
 2022.

\bibitem{Diamanti2015}
O. Diamanti, C. Barnes, S. Paris, E. Shechtman, and O. Sorkine-Hornung,
``Synthesis of complex image appearance from limited exemplars,''
\textit{ACM Transactions on Graphics},
vol.~34, no.~2, pp.~22:1--22:14, 2015.

\bibitem{Donovan2011}
O. Donovan, A. Agarwala, and A. Hertzmann,
``Color compatibility from large datasets,''
\textit{ACM Transactions on Graphics},
vol.~30, no.~4, pp.~63:1--63:12, 2011.

\bibitem{Drago2003}
F. Drago, K. Myszkowski, T. Annen, and N. Chiba,
``Adaptive logarithmic mapping for displaying high contrast scenes,''
\textit{Computer Graphics Forum},
vol.~22, no.~3, pp.~419--426, 2003.

\bibitem{Durand2002}
F. Durand, and J. Dorsey,
``Fast bilateral filtering for the display of high-dynamic-
range images,''
\textit{ACM Transactions on Graphics},
vol.~21, no.~3, pp.~257--266, 2002.



\bibitem{Fan2022}
K. Fan, and S. Liu, 
``A caricature-style generating method for portrait photo,''
\textit{Journal of Computer-Aided Design \& Computer Graphics},
vol. 34, no. 1, pp. 1-8, 2022.

\bibitem{Fattal2002}
R. Fattal, D. Lischinski, and M. Werman,
``Gradient domain high dynamic range compression,''
\textit{ACM Transactions on Graphics},
vol.~21, no.~3, pp.~249--256, 2002.

\bibitem{Gao2021}
R. Gao, Z. Huang, and S. Liu,
``Multi-task deep learning for no-reference screen content image quality assessment,''
In \textit{Proceedings of the 27th International Conference on Multimedia Modeling (MMM)},
Prague, Czech Republic, pp. 213--226, 2021. 

\bibitem{Gao2022}
R. Gao, Z. Huang, and S. Liu,
,''QL-IQA: Learning distance distribution from quality levels for blind image quality assessment,''
\textit{Signal Processing: Image Communication},
vol. 101, 116576: 1--11, 2022.

\bibitem{Hathaway2001}
R.J. Hathaway, and J.C. Bezdek,
``Fuzzy C-means clustering of incomplete data,''
\textit{IEEE Transactions on Systems, Man, and Cybernetics, Part B: Cybernetics},
vol.~31, no.~5, pp.~735--744, 2001.

\bibitem{Hertzmann2001}
A. Hertzmann, C.E. Jacobs, N. Oliver, et al.,
``Image analogies,''
In \textit{Proceedings of ACM SIGGRAPH},
pp.~327--340, 2001.

\bibitem{Efros2015}
A.A. Efros, and W.T. Freeman,
``Image quilting for texture synthesis and transfer,''
In \textit{Proceedings of ACM SIGGRAPH},
pp.~341--346, 2015.

\bibitem{Farbman2008}
Z. Farbman, R. Fattal, D. Lischinski, and R. Szeliski,
``Edge-preserving decompositions for multi-scale tone and detail manipulation,''
\textit{ACM Transactions on Graphics},
vol.~27, no.~3, pp.~67:1--67:10, 2008.

\bibitem{Frigo2019}
 O. Frigo, N. Sabater, J. Delon, and P. Hellier, 
``Video style transfer by consistent adaptive patch sampling,''
\textit{The Visual Computer}, 
vol. 35, no. 3, pp. 429-–443, 2019.

\bibitem{Gao2018}
 C. Gao, D. Gu, F. Zhang, and Y. Yu, 
``ReCoNet: Real-time coherent video style transfer network,''
In \textit{Proceedings of Asian Conference on Computer Vision (ACCV)},
pp. 637-–653, 2018.

\bibitem{Gao2020}
 W. Gao, Y. Li, Y. Yin, and M.-H. Yang, 
``Fast video multi-style transfer,''
In \textit{Proceedings of IEEE Winter Conference on Applications of Computer Vision (WACV)},
pp. 3211–-3219, 2020.

\bibitem{Gatys2016}
L.A. Gatys, A.S. Ecker, and M. Bethge, 
``Image style transfer using convolutional neural networks,''
In \textit{Proceedings of IEEE Conference on Computer Vision and Pattern Recognition (CVPR)}, 
pp. 2414–-2423, 2016.

\bibitem{Gatys2017}
L. Gatys, A. Ecker, M. Bethge, A. Hertzmann, and E. Shechtman,
``Controlling perceptual factors in neural style
transfer,'' 
In \textit{Proceedings of IEEE Conference on Computer Vision and Pattern Recognition (CVPR)},
pp. 3730-–3738, 2017.

\bibitem{Hacohen2011}
Y. Hacohen, E. Shechtman, D.B., Goldman, et al.,
``Non-rigid dense correspondence with applications for image enhancement,''
\textit{ACM Transactions on Graphics},
vol.~30, no.~4, pp.~76--79, 2011.

\bibitem{He2017}
M. He, J. Liao, L. Yuan, and P.V. Sander,
``Neural color transfer between images,''
In \textit{Proceedings of IEEE Conference on Computer Vision and Pattern Recognition},
pp.~1--14, 2017.

\bibitem{He2015}
L. He, H. Qi, and R. Zaretzki,
``Image color transfer to evoke different emotions based on color  combinations,''
\textit{Signal, Image and Video Processing},
vol.~9, no.~8, pp.~1965--1973, 2015.

\bibitem{Huang2017}
X. Huang, and S. Belongie, 
``Arbitrary style transfer in real-time with adaptive instance
normalization,''
In \textit{Proceedings of IEEE International Conference on Computer Vision (ICCV)},
pp. 1510-–1519, 2017.

\bibitem{Huang2017Video}
H. Huang, H. Wang, W. Luo, L. Ma, W. Jiang, X. Zhu, Z. Li, and W. Liu, 
``Real-time neural style transfer for videos,''
In \textit{Proceedings of IEEE Conference on Computer Vision and Pattern Recognition (CVPR)}, pp. 7044-–7052, 2017.


\bibitem{Huang2018MM}
Z. Huang, and S. Liu,
``Robustness and discrimination oriented hashing combining texture and invariant vector distance,''
In \textit{Proceedings of ACM Multimedia},
pp. 1389--1397, 2018.

\bibitem{Hwang2014}
Y. Hwang, J.Y. Lee, I.S. Kweon, et al.,
``Color transfer using probabilistic moving least squares,''
In \textit{Proceedings of IEEE Conference on Computer Vision and Pattern Recognition},
pp.~3342--3349, 2014. 

\bibitem{Irony2005} 
R. Irony, D. Cohen-Or, and D. Lischinski,
``Colorization by example,''
In \textit{Proceedings of Eurographics Symposium on Rendering Techniques}, pp. 201--210, 2005.

\bibitem{Jing2018}
Y. Jing, Y. Liu, Y. Yang, et al.,
``Stroke controllable fast style transfer with adaptive
receptive fields,''
In \textit{Proceedings of European Conference on Computer Vision (ECCV)}, 
pp. 244-–260, 2018.

\bibitem{Johnson1993}
D.R. Jones, C.D. Perttunen, and B.E. Stuckman,
``Lipschitzian optimization without the lipschitz constant,''
\textit{Journal of Optimization Theory and Applications},
vol.~79, no.~1, pp.~157--181, 1993.

\bibitem{Johnson2010}
M.K. Johnson, K. Dale, S. Avidan S, et al.,
``CG2Real: Improving the realism of computer generated images using a large collection of photographs,''
\textit{IEEE Transactions on Visualization and Computer Graphics},
vol.~17, no.~9, pp.~1273--1285, 2010.

\bibitem{Johnson2016} 
J. Johnson, A. Alahi, and F.F. Li, 
``Perceptual losses for real-time style
transfer and super-resolution,''
In \textit{Proceedings of European Conference on Computer Vision (ECCV)},
pp. 694–-711, 2016.

\bibitem{Kobayashi1995}
S. Kobayashi,
``\textit{Art of Color Combinations},''
Kodansha International, Tokyo, 1995.

\bibitem{Kotovenko2019}
D. Kotovenko, A. Sanakoyeu, P. Ma, et al.,
``A content transformation block for image style transfer,'' 
In \textit{Proceedings of IEEE Conference on Computer Vision and Pattern Recognition (CVPR)},
pp. 10032-10041, 2019.

\bibitem{Kozlovtsev2019}
 K. Kozlovtsev, and V. Kitov,
 ``Depth-preserving real-time arbitrary style transfer,''
arXiv preprint: arXiv.1906.01123, 2019

\bibitem{Krizhevsky2017} 
A. Krizhevsky, I. Sutskever, and  G.E. Hinton,
``ImageNET classification with deep convolutional neural networks,''
\textit{Communications of the ACM},
vol. 60, no. 6, pp. 84--90, 2017.

\bibitem{Itten1974}
J. Itten,
``\textit{The Art of Color: the Subjective Experience and Objective Rationale of Color},''
Wiley, New York, 1974.

\bibitem{Lee2016} 
J. Lee, and S. Lee, 
``Hallucination from noon to night images using CNN,''
In \textit{Proceedings of SIGGRAPH Asia Posters}, Article no. 15, 2016.

\bibitem{Levin2004}
A. Levin, D. Lischinski, and Y. Weiss,
``Colorization using optimization,''
\textit{ACM Transactions on Graphics},
vol.~23, no.~3, pp.~689--694, 2004.

\bibitem{Li2005}
Y. Li, L. Sharan, and E. Adelson,
``Compressing and companding high dynamic range images with subband architectures,''
\textit{ACM Transactions on Graphics},
vol.~24, no.~3, pp.~836--844, 2005.

\bibitem{Li2016}
C. Li, and M. Wand,
``Precomputed real-time texture synthesis with markovian generative adversarial networks,''
In \textit{Proceedings of European Conference on Computer Vision (ECCV)}, 
pp. 702--716, 2016.


\bibitem{Li2017}
Y. Li, C. Fang, J. Yang, Z. Wang, X. Lu, and M.-H. Yang,
``Diversified texture synthesis with feed-forward networks,''
In \textit{Proceedings of IEEE Conference on Computer Vision and Pattern Recognition (CVPR)},
pp. 266--274, 2017.

\bibitem{Li2017Universal}
Y. Li, C. Fang, J. Yang, et al.,
``Universal style transfer via feature transforms,''
In \textit{Proceedings of Advances in Neural Information Processing Systems (NIPS)}, 
pp. 386–-396, 2017.

\bibitem{Liao2017}
J. Liao, Y. Yao, L. Yuan, G. Hua, and S.B. Kang, 2017.
``Visual attribute transfer through deep image analogy,''
\textit{arXiv preprint arXiv:1705.01088}, 2017.


\bibitem{Lischinski2006}
D. Lischinski, Z. Farbman, and M. Uyttendaele, 
``Interactive local adjustment of tonal values,''
\textit{ACM Transactions on Graphics},
vol.~25, no.~3, pp.~646--653, 2006.

\bibitem{Liu2017Trcollage}
S. Liu, X. Wang, P. Li, and J. Noh,
``Trcollage: efficient image collage using tree-
based layer reordering,''
In \textit{Proceedings of International Conference on Virtual Reality and Visualization (ICVRV)},
 pp. 454--455, 2017.


\bibitem{Liu2016}
S. Liu, and H. Luo,
``Hierarchical emotional color theme extraction,''
\textit{Color Research \& Application},
vol.~41, no.~5, pp.~513--522, 2016.

\bibitem{Liu2018}
D. Liu, Y. Jiang, M. Pei, and S. Liu,
``Emotional image color transfer via deep learning,''
\textit{Pattern Recognition Letters},
vol.~110, pp.~16--22, 2018.

\bibitem{Liu2019Stitching}
S. Liu, and Q. Chai,
``Shape-optimizing and illumination-smoothing image stitching,''
\textit{IEEE Transactions on Multimedia},
vol. 21, no. 3, pp. 690--703, 2019.

\bibitem{Liu2012}
S. Liu, H. Sun, and X. Zhang,
``Selective color transferring via ellipsoid color mixture map,''
\textit{Journal of Visual Communication and Image Representation},
vol.~23, no.~1, pp.~173--181, 2012.

\bibitem{Liu2018theme}
S. Liu, Y. Jiang, and H. Luo,
``Attention-aware color theme extraction for fabric images,''
\textit{Textile Research Journal},
vol.~88, no.~5, pp.~552--565, 2018.

\bibitem{Liu2019TOMM}
S. Liu, and Z. Huang,
``Efficient image hashing with invariant vector distance for copy detection,''
\textit{ACM Transactions on Multimedia Computing, Communications, and Applications},
vol. 15, no. 4, pp. 106:1--106:22, 2019. 

\bibitem{Liu2018Gamut}
S. Liu, and S. Li,
``Gamut mapping optimization algorithm based on gamut-mapped image measure (GMIM),''
\textit{Signal, Image and Video Processing}, 
vol. 12, no. 1, pp. 67--74, 2018.


\bibitem{Liu2018Access}
S. Liu, and M Pei,
``Texture-aware emotional color transfer between images,''
\textit{IEEE Access},
vol.~6, pp.~31375--31386, 2018.

\bibitem{Liu2019}
S. Liu, Z. Song, X. Zhang, and T. Zhu,
``Progressive complex illumination image appearance transfer based on CNN,'' 
\textit{Journal of Visual Communication and Image Representation}, vol.~64, pp.~102636:1-102636:11, 2019.

\bibitem{Liu2011Toning}
S. Liu, X. Wang, and Q. Peng, 
``Multi-toning image adjustment,''
\textit{Computer Aided Drafting, Design and Manufacturing},
vol.~21, no.~2, pp.~62--72, 2011.

\bibitem{Liu2018AppearanceTansfer}
S. Liu, and Z. Song, 
``Multi-source image appearance transfer based on edit propagation,''
\textit{Journal of Zhengzhou University (Engineering Science)}, vol. 39, no. 5, pp. 22--27, 2018.


\bibitem{Liu2018ICPR}
S. Liu, and Y. Zhang,
``Non-uniform illumination video enhancement based on zone system and fusion,''
In \textit{Proceedings of International Conference on Pattern Recognition (ICPR)}, 
pp. 2711--2716, 2018. 

\bibitem{Liu2019enhancement}
S. Liu, and Y. Zhang,
``Detail-preserving underexposed image enhancement via optimal weighted multi-exposure fusion,'' 
\textit{IEEE Transactions on Consumer Electronics}, Vol. 65, no. 3, pp. 303--311, 2019.

\bibitem{Liu2022}
S. Liu, H. Wang, and M. Pei,
``Facial-expression-aware emotional color transfer based
on convolutional neural network,'' 
\textit{ACM Transactions on Multimedia Computing, Communications, and Applications}, vol.~18, no. 1, 8:1--8:19, 2022.

\bibitem{Liu2021VideoTransfer}
S. Liu, and Y. Zhang,
``Temporal-consistency-aware video color transfer,''
In \textit{Proceedings of Computer Graphics International (CGI)}, pp.~464--476, 2021.

\bibitem{Liu2022MTAP}
S. Liu, and Z. Liu,
``Style enhanced line drawings based on multi-feature,''
\textit{Multimedia Tools and Applications},
vol. 81, 2022.

\bibitem{Liu2022FCS}
S. Liu, and Z. Liu,
``Line drawing via saliency map and ETF,''
\textit{Frontiers of Computer Science},
vol. 16, no. 5, pp. 165707: 1--13, 2022.

\bibitem{Liu2022Style}
S. Liu, and T. Zhu,
``Structure-guided arbitrary style transfer for artistic image and video,''
\textit{IEEE Transactions on Multimedia},
vol. 24, pp. 1299--1312, 2022.

\bibitem{Luan2007}
Q. Luan, F. Wen, and Y. Xu,
``Color transfer brush,''
In \textit{Proceedings of the 15th Pactific Conference on Computer Graphics and Applications},
pp.~465--468, 2007.

\bibitem{Luo2014}
H. Luo, and S. Liu,
``Textile image segmentation through region action graph and novel region merging strategy,''
In \textit{Proceedings of International Conference on Virtual Reality and Visualization (ICVRV)},
2014.


\bibitem{Machajdik2010}
J. Machajdik, and A. Hanbury,
``Affective image classification using features inspired by psychology and art theory,''
In \textit{Proceedings of the International Conference on Multimedia},
pp.~83--92, 2010.

\bibitem{Matsuda1995}
Y. Matsuda,
``\textit{Color Design},''
Asakura Shoten, Tokyo, 1995 (in Japanese).

\bibitem{Means1999}
T.K. Means, S. Wang, and E. Lien,
``Human toll-like receptors mediate cellular activation by mycobacterium tuberculosis,''
\textit{Computer Graphics Forum},
vol.~163, no.~7, pp.~3920--3927, 1999.

\bibitem{Morse2007}
B.S. Morse, D. Thornton, and Q. Xia,
``Image-based color schemes,''
In \textit{Proceedings of IEEE International Conference on Image Processing},
pp.~497--500, 2007.

\bibitem{Nguyen2014}
R.M.H. Nguyen, S.J. Kim, and M.S. Brown,
``Illuminant aware gamut-based color transfer,''
\textit{Computer Graphics Forum},
vol.~33, no.~7, pp.~319--328, 2014.

\bibitem{Okura2015}
F. Okura, K. Vanhoey, A. Bousseau, A.A. Efros, and G. Drettakis, ``Unifying color and texture transfer for predictive appearance manipulation,``
\textit{Computer Graphics Forum},
vol.~34, no.~4, pp.~53--63, 2015.

\bibitem{Park2019}
D. Park, and K. Lee, 
``Arbitrary style transfer with style-attentional networkk,``
In \textit{Proceedings of IEEE Conference on Computer Vision and Pattern Recognition (CVPR)}, 
pp. 5880–-5888, 2019.

\bibitem{Pei2018}
M. Pei, S. Liu, and X. Zhang,
``Facial-expression-aware emotional color transfer based on convolutional neural network,''
In \textit{Proceedings of the 26th Pacific Conference on Computer Graphics and Applications (PG) posters},
pp.~7--8, 2018.

\bibitem{Peng2015}
K.C. Peng, T. Chen, A. Sadovnik, and A. Gallagheret,
``A mixed bag of emotions: Model, predict, and transfer emotion distributions,''
In \textit{Proceedings of IEEE Computer Vision and Pattern Recognition},
pp.~860--868, 2015.

\bibitem{Pitie2007}
F. Piti\'{e}, A.C. Kokaram, and R. Dahyot,
``Automated colour grading using colour distribution transfer,''
\textit{Computer Vision \& Image Understanding},
vol.~107, no.~1--2, pp.~123--137, 2007.

\bibitem{Pouli2011}
T. Pouli, and E. Reinhard,
``Progressive color transfer for images of arbitrary dynamic range,''
\textit{Computers \& Graphics},
vol.~35, pp.~67--80, 2011.

\bibitem{Reinhard2001}
E. Reinhard, M. Ashikhmin, B. Gooch, and P. Shirley,
``Color transfer between images,''
\textit{IEEE Computer Graphics \& Applications},
vol.~21, no.~5, pp.~34--41, 2001.

\bibitem{Reinhard2002}
E. Reinhard, M. Stark, P. Shirley, and J. Ferwerda,
``Photographic tone reproduction for digital images,''
\textit{ACM Transactions on Graphics},
vol.~21, no.~3, pp.~267--276, 2002.

\bibitem{Ruder2016}
M. Ruder, A. Dosovitskiy, and T. Brox,
``Artistic style transfer for videos,''
In \textit{Proceedings of German Conference on Pattern Recognition}, pp. 26–-36, 2016.

\bibitem{Ruder2018}
M. Ruder, A. Dosovitskiy, and T. Brox,
``Artistic style transfer for videos and spherical
images,'' 
\textit{International Journal of Computer Vision}, 
vol. 126, no. 11, pp. 1199-–1219, 2018.

\bibitem{Sanakoyeu2018}
A. Sanakoyeu, D. Kotovenko, S. Lang, and B. Ommer,
``A style-aware content loss for realtime HD style transfer,''
In \textit{Proceedings of European Conference on Computer Vision (ECCV)},
pp. 715-–731, 2018.

\bibitem{Sener2012}
O. Sener, K. Ugur, and A.A. Alatan,
``Error-tolerant interactive image segmentation using dynamic and iterated graph-cuts,''
In \textit{Proceedings of the 2nd ACM International Workshop on Interactive Multimedia on Mobile and Portable Devices},
pp.~9--16, 2012.

\bibitem{Shen2018}
F. Shen, S. Yan, and G. Zeng,
``Neural style transfer via meta networks,''
In \textit{Proceedings of IEEE Conference on Computer Vision and Pattern Recognition (CVPR)},
pp. 8061–8069, 2018.

\bibitem{Shih2013}
Y. Shih, S. Paris, F. Durand, and W.T. Freeman,
``Data-driven hallucination of different times of day from a single outdoor photo,''
\textit{ACM Transactions on Graphics},
vol.~32, no.~6, pp.~2504--2507, 2013.


\bibitem{Solli2010}
M. Solli, and R. Lenz,
``Emotion related structures in large image databases,''
In \textit{Proceedings of the ACM International Conference on Image and Video Retrieval},
pp.~398--405, 2010.

\bibitem{Song2017}
Z. Song, and S Liu,
``Sufficient image appearance transfer combining color and texture,'' 
\textit{IEEE Transactions on Multimedia}, vol.~19, no.~4, pp.~702--711, 2017.


\bibitem{Subr2013}
K. Subr, S. Paris, C. Soler, and J. Kautz J,
``Accurate binary image selection from inaccurate user input,''
\textit{Computer Graphics Forum},
vol.~32, no.~2, pp.~41--50, 2013.

\bibitem{Tai2007}
Y.W. Tai, J. Jia, and C.K. Tang,
``Soft color segmentation and its applications,''
\textit{IEEE Transactions on Pattern Analysis \& Machine Intelligence},
vol.~29, no.~9, pp.~1520--1537, 2007.

\bibitem{Terry2022}
https://petapixel.com/2022/01/12/this-web-app-can-use-one-photos-colors-to-grade-another/

\bibitem{Terry2022CT}
https://www.dustfreesolutions.com/CT/CT.html

\bibitem{Tibshirani1996}
R. Tibshirani,
``Regression shrinkage and selection via the lasso,''
\textit{Journal of the Royal Statistical Society Series B (Methodological)},
vol.~58, no.~1, pp.~267--288, 1996.

\bibitem{Ulyanov2016}
D. Ulyanov, V. Lebedev, A. Vedaldi, and V.S. Lempitsky,
``Texture networks:
Feed-forward synthesis of textures and stylized images,''
In \textit{Proceedings of the 33rd International Conference on International Conference on Machine Learning (ICML)}, 
pp. 1349-–1357, 2016.

\bibitem{Ulyanov2017}
 D. Ulyanov, A. Vedaldi, and V. Lempitsky,
``Improved texture networks: maximizing
quality and diversity in feed-forward stylization and texture synthesis,'' 
In \textit{Proceedings of IEEE
Conference on Computer Vision and Pattern Recognition}, 
pp. 4105--4113, 2017.

\bibitem{Wang2004}
C.M. Wang, and Y.H. Huang,
``A novel color transfer algorithm for image sequences,''
\textit{Journal of Information Science and Engineering},
vol.~20, no.~6, pp.~1039--1056, 2004.

\bibitem{Wang2006}
C.M. Wang, Y.H. Huang, and M.L. Huang,
``An effective algorithm for image sequence color transfer,''
\textit{Mathematical and Computer Modelling},
vol.~44, no.~7, pp.~608--627, 2006.

\bibitem{Wang2013}
X. Wang, J. Jia, and L. Cai,
``Affective image adjustment with a single word,''
\textit{The Visual Computer},
vol.~29, no.~11, pp.~1121--1133, 2013.

\bibitem{Wang2017}
X. Wang, G. Oxholm, D. Zhang, and Y.-F. Wang,
``Multimodal transfer: a hierarchical deep convolutional
neural network for fast artistic style transfer,''
In \textit{Proceedings of IEEE Conference
on Computer Vision and Pattern Recognition (CVPR)},
pp. 7178-–7186, 2017.

\bibitem{Ward1997}
 G. Ward, H. Rushmeier, and C. Piatko,
``A visibility matching tone reproduction operator for high dynamic range scenes,''
\textit{IEEE Transactions on Visualization and Computer Graphics},
vol.~3, no.~4, pp.~291--306, 1997.

\bibitem{Wei2009}
L. Wei, S. Lefebvre, V. Kwatra, and G. Turk,
``state of the art in example-based texture synthesis,''
In \textit{Proceedings of Eurographics 2009, State of the Art Report, EG-STAR},
pp.~1--25, 2009.

\bibitem{Wen2008}
C. Wen, H. Chang-Hsi, B. Chen B, et al.,
``Example-based multiple local color transfer by strokes,''
\textit{Computer Graphics Forum},
vol.~27, no.~7, pp.~1765--1772, 2008.

\bibitem{Xiao2006}
X. Xiao, and L. Ma,
``Color transfer in correlated color space,''
In \textit{Proceedings of ACM International Conference on Virtual Reality Continuum and Its Applications},
pp.~305--309, 2006.

\bibitem{Xu2019}
Z. Xu, M.J. Wilber, C. Fang C, et al.,
``Learning from multi-domain artistic images for
arbitrary style transfer,''
In \textit{Proceedings of Eurographics Expressive Symposium}, 
pp. 21-–31, 2019.



\bibitem{Yang2008}
C.K. Yang, and L.K. Peng,
``Automatic mood-transferring between color images,''
\textit{IEEE Computer Graphics \& Applications},
vol.~28, no.~2, pp.~52--61, 2008.

\bibitem{Yao2019}
Y. Yao, J. Ren, X. Xie, W. Liu, Y. Liu, and J. Wang,
``Attention-aware multi-stroke style transfer,''
In \textit{Proceedings of IEEE Conference on Computer Vision and Pattern Recognition (CVPR)},
 pp. 1467-–1475, 2019.

\bibitem{Zhang2011}
X. Zhang, and S. Liu,
``Texture transfer in frequency domain,''
In \textit{Proceedings of the International Conference on Image and Graphics (ICIG)},
pp.~123--128, 2011.

\bibitem{Zhang2018}
H. Zhang, and K. Dana, 
``Multi-style generative network for real-time transfer,''
In \textit{Proceedings of European Conference on Computer Vision (ECCV) Workshops}, 
pp. 349-–365, 2018.

\bibitem{Zhang2020Access}
J. Zhang, S. Liu, J. Fan, J. Huang, and M. Pei, ``NK-CDS: A creative design system for museum art derivatives,''
\textit{IEEE Access},
vol. 8, no. 1, pp. 29259--29269, 2020.


\bibitem{Zhang2016Asia}
Y. Zhang, and S. Liu,
``Effective underexposed video enhancement via optimal fusion,''
In \textit{Proceedings of the 9th ACM SIGGRAPH Conference and Exhibition on Computer Graphics and Interactive Techniques in Asia (SIGGRAPH Asia 2016), poster}, 
Article No. 16, 2016.

\bibitem{Zhang2017CAD}
Y. Zhang, and S. Liu, 
``Non-uniform illumination video enhancement based on zone system and fusion,''
\textit{Journal of Computer-Aided Design \& Computer Graphics}, vol.~29, no.~12, pp.~2317--2322, 2017.

\bibitem{Zhang2019Heritage}
J. Zhang, Z. Liu, and S. Liu,
``Computer simulation of archaeological drawings,'' \textit{Journal of Cultural Heritage},
vol. 37, pp. 181--191, 2019. 

\bibitem{Zheng2012}
D. Zheng, Y. Han, and G. Baciu,
``Design through cognitive color theme: A new approach for fabric design,''
In \textit{Proceedings of IEEE 11th International Conference on Cognitive Informatics \& Cognitive Computing},
pp.~346--355, 2012.

\bibitem{Zhu2020}
T. Zhu, and S. Liu,
``Detail-preserving arbitrary style transfer,''
In \textit{Proceedings of the IEEE International Conference on Multimedia and Expo (ICME)},
pp. 1--6, 2020. 

\end{thebibliography}
\end{document}